\documentclass[conference]{IEEEtran}
\IEEEoverridecommandlockouts
\usepackage{cite}
\usepackage{amsmath,amssymb,amsfonts}
\usepackage{algorithmic}
\usepackage{graphicx}
\usepackage{textcomp}
\usepackage{xcolor}
\usepackage{float}
\usepackage{booktabs}
\def\BibTeX{{\rm B\kern-.05em{\sc i\kern-.025em b}\kern-.08em
    T\kern-.1667em\lower.7ex\hbox{E}\kern-.125emX}}
\begin{document}

\title{KIO-planner: Attention-Guided Single-Stage Motion Planning with Dual Mapping for UAV Navigation%
\thanks{This work has been accepted for publication in an IEEE Vehicular Technology Conference (VTC). \copyright~2026 IEEE. Personal use of this material is permitted. Permission from IEEE must be obtained for all other uses, in any current or future media, including reprinting/republishing this material for advertising or promotional purposes, creating new collective works, for resale or redistribution to servers or lists, or reuse of any copyrighted component of this work in other works.}}

\author{\IEEEauthorblockN{1\textsuperscript{st} Dexing Yao†}
\IEEEauthorblockA{\textit{Faculty of Applied Sciences} \\
\textit{Macao Polytechnic University}\\
Macao, China \\
p2522978@mpu.edu.mo}
\and
\IEEEauthorblockN{1\textsuperscript{st} Haochen Li†}
\IEEEauthorblockA{\textit{Faculty of Applied Sciences} \\
\textit{Macao Polytechnic University}\\
Macao, China \\
p2523372@mpu.edu.mo}
\and
\IEEEauthorblockN{2\textsuperscript{nd} Junhao Wei}
\IEEEauthorblockA{\textit{Faculty of Applied Sciences} \\
\textit{Macao Polytechnic University}\\
Macao, China \\
p2312195@mpu.edu.mo}
\and
\IEEEauthorblockN{3\textsuperscript{rd} Yifu Zhao}
\IEEEauthorblockA{\textit{Faculty of Applied Sciences} \\
\textit{Macao Polytechnic University}\\
Macao, China \\
p2523269@mpu.edu.mo}
\and
\IEEEauthorblockN{4\textsuperscript{th} Yanxiao Li}
\IEEEauthorblockA{\textit{Faculty of Applied Sciences} \\
\textit{Macao Polytechnic University}\\
Macao, China \\
p2525981@mpu.edu.mo}
\and
\IEEEauthorblockN{5\textsuperscript{th} Jiahui Xu}
\IEEEauthorblockA{\textit{Faculty of Applied Sciences} \\
\textit{Macao Polytechnic University}\\
Macao, China \\
p2525956@mpu.edu.mo}
\and
\IEEEauthorblockN{6\textsuperscript{th} Jinxuan Hu}
\IEEEauthorblockA{\textit{Faculty of Applied Sciences} \\
\textit{Macao Polytechnic University}\\
Macao, China \\
p2514566@mpu.edu.mo}

\and
\IEEEauthorblockN{7\textsuperscript{th} Lele Tian}
\IEEEauthorblockA{\textit{Faculty of Applied Sciences} \\
\textit{Macao Polytechnic University}\\
Macao, China \\
p2314709@mpu.edu.mo}
\and
\IEEEauthorblockN{8\textsuperscript{th} Baili Lu}
\IEEEauthorblockA{\textit{College of Animal Science and Technology} \\
\textit{Zhongkai University of Agriculture and Engineering}\\
Guangzhou, China \\
18023304003@163.com}
\and
\IEEEauthorblockN{9\textsuperscript{th} Zikun Li}
\IEEEauthorblockA{\textit{School of Economics and Management} \\
\textit{South China Normal University}\\
Macao, China \\
20190731013@m.scnu.edu.cn}
\and
\IEEEauthorblockN{10\textsuperscript{th} Xu Yang}
\IEEEauthorblockA{\textit{Faculty of Applied Sciences} \\
\textit{Macao Polytechnic University}\\
Macao, China \\
xuyang@mpu.edu.mo}

\and
\IEEEauthorblockN{11\textsuperscript{th} Sio-Kei Im}
\IEEEauthorblockA{\textit{Faculty of Applied Sciences} \\
\textit{Macao Polytechnic University}\\
Macao, China \\
marcusim@mpu.edu.mo}
\and

\IEEEauthorblockN{12\textsuperscript{th} Dingcheng Yang}
\IEEEauthorblockA{\textit{Information Engineering School} \\
\textit{Nanchang University}\\
Nanchang, China \\
yangdingcheng@ncu.edu.cn}
\and

\IEEEauthorblockN{13\textsuperscript{th} Yapeng Wang{*}}
\IEEEauthorblockA{\textit{Faculty of Applied Sciences} \\
\textit{Macao Polytechnic University}\\
Macao, China \\
yapengwang@mpu.edu.mo}
{*}Corresponding author
}

\maketitle

\begin{abstract}
Autonomous UAV flight in confined, wall-dense environments requires low-latency and reliable motion planning under strict safety constraints. Traditional optimization-based planners suffer from mapping latency and easily fall into local minima when navigating through dense structural obstacles. Meanwhile, existing end-to-end learning methods struggle to extract fine-grained geometric features from raw depth images and lack hard kinodynamic constraints, leading to unpredictable collisions near walls. To address these issues, we propose KIO-planner, an attention-guided single-stage trajectory planning framework. First, we integrate a Convolutional Block Attention Module (CBAM) into the perception backbone to adaptively focus on critical structural edges and traversable space. Second, we introduce a novel Dual Mapping mechanism—comprising physical bounds activation and a deterministic Geometric Safety Shield in the depth-pixel space—to enforce kinodynamic feasibility and collision-free flight without global map fusion. Extensive high-fidelity simulated experiments demonstrate that KIO-planner enables highly agile navigation at speeds up to 3.0 m/s. Compared to the state-of-the-art baseline, KIO-planner achieves lower inference latency ($\sim$24 ms) and generates significantly smoother trajectories (reducing control cost by 28.4\%). Most notably, our Dual Mapping substantially increases the worst-case safety margin (minimum distance to obstacles) from 0.48 m to 0.76 m, ensuring fast, smooth, and safer navigation in highly constrained environments.
\end{abstract}


\section{INTRODUCTION}
In recent years, the deployment of Unmanned Aerial Vehicles (UAVs) in confined and structured environments, such as indoor navigation, tight corridors, and industrial facility inspections, has significantly increased. High-speed autonomous navigation in these wall-dense scenarios requires extreme agility, low latency, and reliable safety. Traditional optimization-based planners, such as Fast-Planner [1] and EGO-planner [2], have achieved strong performance by utilizing Euclidean Signed Distance Fields (ESDFs) or collision-free guiding paths to optimize trajectories. Furthermore, advanced trajectory optimization frameworks like MINCO [3] leverage smooth maps to handle geometric constraints efficiently. However, in highly constrained environments with dense uncooperative walls, these decoupled architectures still suffer from significant computational mapping latency and are highly prone to getting trapped in local minima during gradient optimization.

To bypass the latency of explicit mapping and complex optimization, learning-based end-to-end navigation approaches have emerged as a promising alternative. Recent pioneering works have demonstrated the promising potential of convolutional neural networks (CNNs) in mapping sensory observations directly to collision-free trajectories, enabling high-speed flight in the wild [4] and agile dynamic obstacle avoidance [5]. Despite their fast inference speeds, these 'black-box' models suffer from a major limitation: the lack of explicit kinodynamic and geometric hard constraints. When operating in extremely close proximity to walls, even a minor probabilistic prediction error can lead to collision risks.

To overcome these challenges, we propose KIO-planner, an attention-guided single-stage trajectory generation framework equipped with improved safety guarantees. To significantly enhance the network's spatial perception, we seamlessly integrate the Convolutional Block Attention Module (CBAM) \cite{b6} into our feature extraction backbone. By adaptively recalibrating both channel and spatial feature responses, KIO-planner accurately focuses on critical navigable gaps while suppressing irrelevant background noise. 

More importantly, to bridge the gap between neural network unpredictability and hard safety constraints, we introduce a novel Dual Mapping mechanism. Inspired by the concepts of lazy search over local 3D data \cite{b7} and safe learning via shielding \cite{b8}, our Geometric Safety Shield deterministically evaluates and crops unsafe trajectory candidates directly in the depth-pixel space, eliminating the need for global map fusion. Finally, to ensure kinodynamic feasibility, the framework generates closed-form polynomial motion primitives \cite{b9} that are strictly mapped to the UAV's physical actuation limits.

The main contributions of this paper are as follows:
\begin{itemize}
    \item We propose KIO-planner, an efficient single-stage planning framework that achieves millisecond-level inference for high-speed UAV navigation in confined wall environments.
    \item We introduce a CBAM-enhanced perception backbone that significantly improves the network's ability to extract fine-grained structural edges and narrow gaps.
    \item We design a Dual Mapping mechanism, comprising physical bounds activation and a Geometric Safety Shield, to provide a deterministic safety margin and guarantee collision avoidance.
    \item Extensive high-fidelity simulated experiments demonstrate that KIO-planner achieves superior agility, smoothness, and safety compared to state-of-the-art baselines.
\end{itemize}
\begin{figure*}[htbp] 
    \centering
    \includegraphics[width=0.95\textwidth]{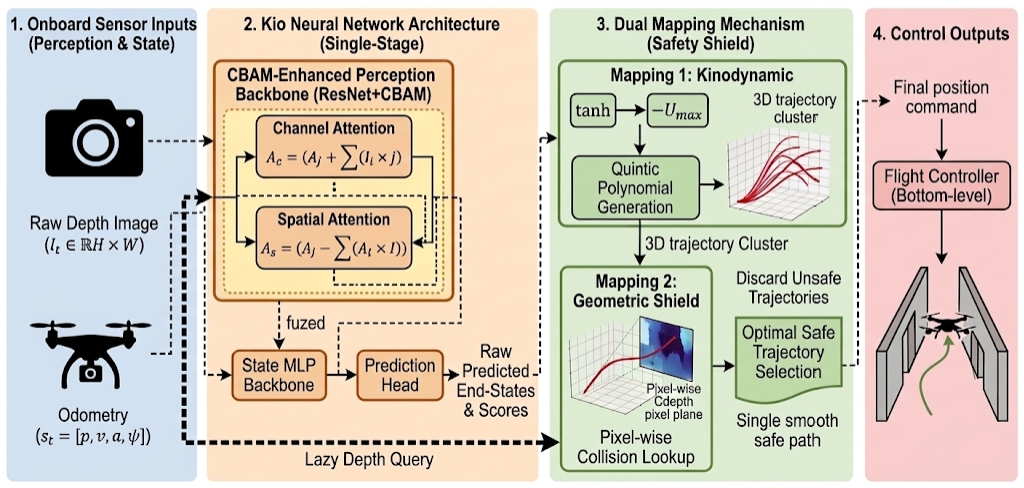} 
    \caption{The proposed KIO-planner framework: The depth image and odometry are processed by a CBAM-enhanced backbone to predict raw primitives, which are then strictly constrained by the Dual Mapping mechanism to ensure kinodynamic and geometric safety.}
    \label{fig:framework}
\end{figure*}

\section{RELATED WORK}

\subsection{UAV Local Motion Planning}
Traditional UAV local planning typically follows a hierarchical 'perception-mapping-planning' pipeline. Gradient-based trajectory optimizers, such as Fast-Planner \cite{b1} and EGO-planner \cite{b2}, have demonstrated impressive agility by utilizing Euclidean Signed Distance Fields (ESDFs) or collision-free guiding paths to push trajectories away from obstacles. To handle complex geometric and kinodynamic constraints, advanced frameworks like MINCO \cite{b3} parameterize trajectories with spatial-temporal deformations. However, in highly confined environments with dense, non-convex walls, these optimization-based methods often suffer from severe mapping latency and easily get trapped in local minima, leading to oscillatory behaviors. 

To mitigate explicit mapping overhead, learning-based end-to-end approaches have gained significant traction. Kaufmann et al. \cite{b4} demonstrated that mapping raw sensory observations directly to actions can enable high-speed flight in the wild. Similarly, Deep-PANTHER \cite{b5} leverages imitation learning for rapid perception-aware replanning. Despite the ultra-low latency of end-to-end neural planners, these 'black-box' models generally lack explicit kinodynamic boundaries. Consequently, their probabilistic predictions often lead to collision risks when navigating near walls where the margin for error is close to zero.

\subsection{Attention-Guided Representation Learning}
Extracting robust geometric features from high-dimensional depth images is crucial for reliable navigation. While standard Convolutional Neural Networks (CNNs) fuse spatial and channel-wise information within local receptive fields, their representational power can be further enhanced by attention mechanisms. SENet \cite{b10} introduced the Squeeze-and-Excitation block to adaptively recalibrate channel-wise feature responses, proving highly effective in classification tasks. However, in UAV navigation, the spatial location of structural edges and narrow gaps is just as critical as the channel features. Therefore, our KIO-planner framework incorporates the Convolutional Block Attention Module (CBAM) \cite{b6}, which sequentially infers attention maps along both channel and spatial dimensions. This dual attention allows KIO-planner to explicitly focus on traversable spaces while ignoring irrelevant background noise, a capability missing in standard learning-based planners.

\subsection{Safety-Critical and Kinodynamic Navigation}
Ensuring absolute safety during high-speed flight requires bridging the gap between motion planning and control limits. Mueller et al. \cite{b9} pioneered the rapid generation of closed-form polynomial motion primitives, establishing a foundation for analytically guaranteeing kinodynamic feasibility. In terms of obstacle avoidance, building and maintaining global maps introduces uncertainties and delays. Florence et al. proposed NanoMap \cite{b7}, demonstrating that querying pose-uncertainty-aware local 3D data via lazy search is far more efficient for fast flight. 

Furthermore, in the domain of Safe Reinforcement Learning, Alshiekh et al. \cite{b8} introduced the concept of 'Shielding' to monitor and correct unsafe actions from a learning agent. Inspired by these principles, KIO-planner introduces a novel Dual Mapping mechanism. It directly maps network outputs to the physical actuation limits and projects generated kinodynamic polynomials onto the 2D depth-pixel plane for a deterministic safety shield lookup. This design completely bypasses global mapping while locking in a hard geometric safety margin. More broadly, recent UAV planning studies have also investigated self-attention and goal-aware anchor representations \cite{b11}, swarm-intelligence-based 2D/3D path planning \cite{b12}, \cite{b13}, adaptive particle-swarm updates \cite{b14}, and bandit hyper-heuristics for UAV inspection routing \cite{b15}. Related optimization and learning-aided studies have further explored enhanced whale-optimization variants \cite{b16}, academic-potential prediction \cite{b17}, stock-price forecasting \cite{b18}, and heuristic RRT-based UAV urban path planning \cite{b19}.

\section{METHODOLOGY}

\subsection{Problem Formulation and System Overview}
The dynamic state of the quadrotor is defined by its flat outputs $\mathbf{s} = [\mathbf{p}, \mathbf{v}, \mathbf{a}, \psi]^T$. The local planning task is formulated as finding a dynamically feasible terminal state $\mathbf{x}_T = [\mathbf{p}_T, \mathbf{v}_T, \mathbf{a}_T]^T \in \mathbb{R}^9$ within a short time horizon $T_f$. At each decision step $t$, the KIO-planner network $\pi_\theta$ directly maps the current depth image $\mathbf{I}_t \in \mathbb{R}^{H \times W}$ and the kinodynamic state $\mathbf{s}_t$ to a set of optimal motion primitives $\mathcal{T}$:
\begin{equation}
    \mathcal{T} = \pi_\theta(\mathbf{I}_t, \mathbf{s}_t) = \{(\mathbf{x}_T^{(k)}, c^{(k)})\}_{k=1}^{K}
\end{equation}
where $c^{(k)} \in (0,1)$ is the predicted confidence score for the $k$-th trajectory. The system completely bypasses explicit volumetric mapping (e.g., ESDF) and operates in a single-stage, end-to-end manner.

\subsection{CBAM-Enhanced Perception Backbone}
To accurately extract critical geometric features such as sharp wall edges and narrow traversable gaps from the high-dimensional depth image, we embed the Convolutional Block Attention Module (CBAM) \cite{b6} into our ResNet-based perception backbone. Given an intermediate feature map $\mathbf{F} \in \mathbb{R}^{C \times H \times W}$, KIO-planner sequentially applies attention mechanisms to refine the features:
\begin{equation}
    \mathbf{F}' = \mathbf{M}_c(\mathbf{F}) \otimes \mathbf{F}, \quad \mathbf{F}'' = \mathbf{M}_s(\mathbf{F}') \otimes \mathbf{F}'
\end{equation}
where $\otimes$ denotes element-wise multiplication.

\textit{1) Channel Attention:} To highlight channels containing severe depth gradients, spatial information is aggregated via average-pooling ($\mathbf{F}_{avg}^c$) and max-pooling ($\mathbf{F}_{max}^c$). The channel attention map $\mathbf{M}_c \in \mathbb{R}^{C \times 1 \times 1}$ is computed as:
\begin{equation}
    \mathbf{M}_c(\mathbf{F}) = \sigma\Big(W_1\big(W_0(\mathbf{F}_{avg}^c)\big) + W_1\big(W_0(\mathbf{F}_{max}^c)\big)\Big)
\end{equation}
where $\sigma$ is the Sigmoid function, and $W_0, W_1$ are shared multi-layer perceptron (MLP) weights.

\textit{2) Spatial Attention:} In dense wall environments, the spatial location of gaps is crucial. A 2D spatial mask $\mathbf{M}_s \in \mathbb{R}^{1 \times H \times W}$ is generated by applying a $7 \times 7$ convolution $f^{7\times 7}$ over the concatenated pooled features along the channel axis, forcing the network to concentrate on the free space directly ahead:
\begin{equation}
    \mathbf{M}_s(\mathbf{F}') = \sigma\Big(f^{7 \times 7}\big([\mathbf{F}_{avg}^s; \mathbf{F}_{max}^s]\big)\Big)
\end{equation}
Furthermore, the sequential arrangement of channel and spatial attention is deliberately chosen to align with the visual characteristics of depth-based obstacle perception in confined environments. While the channel attention first filters out uninformative feature maps to reduce background noise, the subsequent spatial attention explicitly highlights the geometric contours of narrow gaps and sharp wall edges. This hierarchical feature refinement is crucial: it allows our lightweight perception backbone to achieve a highly robust, quasi-global understanding of the traversable space without incurring the prohibitive computational overhead and latency associated with complex self-attention mechanisms (e.g., Vision Transformers).
\subsection{Dual Mapping Mechanism (Safety Shield)}
To strictly guarantee kinodynamic feasibility and geometric safety without global map fusion, we propose a deterministic Dual Mapping mechanism at the network's output.

\textit{1) Mapping 1: Physical Bounds Activation:} The raw kinematic output $\mathbf{h}_{kin}$ from the prediction head is unbounded. To ensure the predicted terminal state adheres to the UAV's actuator limits, we apply a hyperbolic tangent activation scaled by the maximum physical envelope $\mathbf{U}_{max} = \text{diag}([P_{max}, V_{max}, A_{max}])$:
\begin{equation}
    \mathbf{x}_T = \mathbf{U}_{max} \odot \tanh(\mathbf{h}_{kin})
\end{equation}
Based on $\mathbf{x}_T$ and the initial state $\mathbf{x}_0$, a closed-form quintic polynomial trajectory \cite{b9} $\mathbf{p}(t)$ is analytically generated.

\textit{2) Mapping 2: Geometric Safety Shield:} To eliminate probabilistic out-of-distribution (OoD) collision risks near walls, the generated 3D polynomial trajectory is discretized into waypoints $\mathbf{p}_w^{(m)}$. Using the UAV pose $(\mathbf{R}_{wb}, \mathbf{t}_{wb})$ and camera intrinsics $\mathbf{K}$, these points are projected back onto the pixel plane $(u, v)$ of the current depth map $\mathbf{I}_t$:
\begin{equation}
    \begin{bmatrix} u \\ v \\ 1 \end{bmatrix} \sim \mathbf{K} \cdot \mathbf{R}_{bc}^T \mathbf{R}_{wb}^T (\mathbf{p}_w^{(m)} - \mathbf{t}_{wb})
\end{equation}
A candidate trajectory is deterministically rejected if the camera-frame depth $p_{c,z}^{(m)}$ of any waypoint violates the observed metric depth $d_{obs}(u,v)$ by breaching the UAV's safety radius margin $r$:
\begin{equation}
    p_{c,z}^{(m)} > d_{obs}(u,v) - (r + \epsilon)
\end{equation}
This pixel-wise lookup strictly enforces the geometric safety constraints, significantly enhancing collision avoidance robustness at a negligible computational cost.
\subsection{Unsupervised Trajectory Optimization}
KIO-planner is trained in an unsupervised manner using an Optimal Boundary Value Problem (OBVP) formulation. The total loss $\mathcal{L}_{total}$ is balanced by empirical weights and comprises three key components:
\begin{equation}
    \mathcal{L}_{total} = \lambda_s \mathcal{L}_{smooth} + \lambda_c \mathcal{L}_{safety} + \lambda_g \mathcal{L}_{guidance}
    \label{eq:total_loss}
\end{equation}

Specifically, to ensure agile yet smooth flight, we analytically minimize the jerk integral. Given the closed-form polynomial solution, the smoothness cost is elegantly decoupled into a quadratic form of the state difference vector $\mathbf{d}_{\mu} = [\mathbf{x}_{0}^{(\mu)}, \mathbf{x}_{T}^{(\mu)}]^{T}$:
\begin{equation}
    \mathcal{L}_{smooth} = \sum_{\mu \in \{x,y,z\}} \mathbf{d}_{\mu}^{T} \mathbf{R}_J \mathbf{d}_{\mu}
    \label{eq:smoothness}
\end{equation}
where $\mathbf{R}_{J}$ is the predefined analytical jerk penalty matrix. To systematically address the common failure mode of traditional planners—getting trapped in local minima when confronting continuous flat walls—we elegantly decouple the guidance objective. Rather than employing a simple Euclidean distance penalty that often generates contradictory gradients near obstacles, our guidance loss evaluates both parallel progress toward the goal and orthogonal lateral exploration. This strategic relaxation explicitly encourages the network to actively sample and discover adjacent safe gaps along the wall surface. 

To formulate the safety loss $\mathcal{L}_{safety}$ without explicitly building a volumetric map during training, we utilize a margin-based depth penalty. For each waypoint $\mathbf{p}_w^{(m)}$ of the predicted trajectory, we project it onto the depth image to obtain its camera-frame depth $p_{c,z}^{(m)}$ and the corresponding ground-truth observed depth $d_{obs}(u,v)$. The penalty is computed using a Softplus function over the margin violations:
\begin{equation}
    \mathcal{L}_{safety} = \sum_{k=1}^{K} c^{(k)} \sum_{m} \text{Softplus}\left( p_{c,z}^{(m)} - d_{obs}(u,v) + r + \epsilon \right)
    \label{eq:safety_loss}
\end{equation}
where $c^{(k)}$ is the predicted confidence score, $r$ is the UAV physical radius, and $\epsilon$ is a designated safety buffer. This formulation encourages the network to focus its outputs exclusively within the valid free space and strongly suppresses collision-prone primitives early in the training phase.
\section{EXPERIMENTS}

To evaluate the performance of the proposed KIO-planner framework, we conduct extensive simulated experiments in highly confined, wall-dense environments. We compare KIO-planner against a state-of-the-art gradient-based baseline (EGO-planner). Furthermore, to validate the effectiveness of our Dual Mapping mechanism, we introduce an ablation variant, \textit{KIO-planner (w/o Safety Shield)}, which removes the Geometric Safety Shield.

\subsection{Experimental Setup and Environment Complexity}
To rigorously evaluate the kinodynamic limits and robustness of the proposed framework, all comparative experiments are conducted in a high-fidelity simulated environment based on ROS Noetic. The simulated testbed is specifically designed to be extremely challenging, covering a massive $100 \times 100 \times 16$ m spatial domain. To severely test the perception and mapping capabilities of the algorithms, this area is procedurally injected with $300$ randomly generated continuous wall formations. This extreme obstacle density creates a maze-like structure featuring narrow vertical gaps, non-convex blind corners, and highly restricted flight corridors.

The simulated quadrotor is equipped with a forward-facing depth camera, providing raw depth image streams at 30 Hz with a maximum sensing range of 5.0 meters. To accurately reflect the computational constraints of modern UAVs, the network inference and all baseline algorithms are benchmarked using an onboard edge computing profile equivalent to the NVIDIA Jetson Orin NX module.

To progressively push the planning algorithms to their absolute limits, the maximum flight velocity is strictly constrained to three distinct tiers: $v_{max} \in \{2.0, 2.5, 3.0\}$ m/s. For each method and speed setting, we execute multiple independent simulated flight trials with heavily randomized start and goal configurations. The quantitative results reported in our evaluation (e.g., Latency, Minimum Distance) represent the average values calculated across these randomized runs to ensure reliable evaluation.

\begin{table*}[htbp]
\caption{Comprehensive quantitative evaluation in wall-dense environments (averaged over multiple randomized trials)}
\centering
\resizebox{\textwidth}{!}{
\begin{tabular}{l l c c c c c c}
\toprule
\textbf{Speed Limit} & \textbf{Method} & \textbf{Latency (ms)} $\downarrow$ & \textbf{Path Length (m)} $\downarrow$ & \textbf{Avg. Speed (m/s)} $\uparrow$ & \textbf{Max. Speed (m/s)} & \textbf{Min. Dist. (m)} $\uparrow$ & \textbf{Smoothness} $\downarrow$ \\
\midrule

& EGO-planner & 38.62 & 56.73 & 1.47 & 2.46 & 0.65 & 2.47 \\
$v_{max} = 2.0$ m/s & KIO-planner (w/o Safety Shield) & 36.52 & 54.42 & 1.49 & 2.09 & 0.62 & 2.17 \\
& \textbf{KIO-planner (Ours)} & \textbf{35.99} & \textbf{54.06} & \textbf{1.51} & \textbf{2.09} & \textbf{0.71} & \textbf{1.73} \\

\midrule

& EGO-planner & 30.66 & 56.88 & 1.86 & 2.96 & 0.46 & 8.13 \\
$v_{max} = 2.5$ m/s & KIO-planner (w/o Safety Shield) & 30.63 & 55.21 & 1.80 & 2.63 & 0.55 & 7.72 \\
& \textbf{KIO-planner (Ours)} & \textbf{27.29} & \textbf{53.60} & \textbf{1.97} & \textbf{2.62} & \textbf{0.86} & \textbf{4.94} \\

\midrule

& EGO-planner & 26.08 & 56.60 & 2.19 & 3.64 & 0.48 & 19.00 \\
$v_{max} = 3.0$ m/s & KIO-planner (w/o Safety Shield) & 25.02 & 55.06 & 2.21 & 3.19 & 0.65 & 15.87 \\
& \textbf{KIO-planner (Ours)} & \textbf{23.99} & \textbf{54.14} & \textbf{2.27} & \textbf{3.18} & \textbf{0.76} & \textbf{13.59} \\

\bottomrule
\end{tabular}
}
\label{tab:quantitative_full}
\end{table*}

\subsection{Quantitative Baseline Comparison}
As summarized in Table \ref{tab:quantitative_full}, the proposed KIO-planner framework significantly outperforms the EGO-planner baseline across all critical metrics, especially under aggressive speed limits. At $v_{max} = 3.0$ m/s, the baseline method exhibits a dangerous minimum safety distance of $0.48$ m, cutting corners too closely and risking catastrophic collisions. In stark contrast, KIO-planner expands the worst-case safety margin to $0.76$ m (a $58.3\%$ improvement) while maintaining a higher average flight speed of $2.27$ m/s. 

Furthermore, the integration of CBAM drastically enhances the perception of navigable space, allowing KIO-planner to generate much smoother control commands. The smoothness cost drops significantly from $19.00$ (EGO-planner) to $13.59$ (KIO-planner). Although the single-stage architecture shifts the computational burden from iterative optimization to neural forward passes, the integration of the lightweight CBAM and the depth-pixel safety shield ensures that KIO-planner maintains an ultra-low inference latency of $\sim 24$ ms. This efficiency intrinsically avoids the mapping delays typical in optimization-based pipelines.

\subsection{Ablation Study and Qualitative Trajectory Analysis}
\begin{figure}[htbp]
    \centering
    \includegraphics[width=\columnwidth]{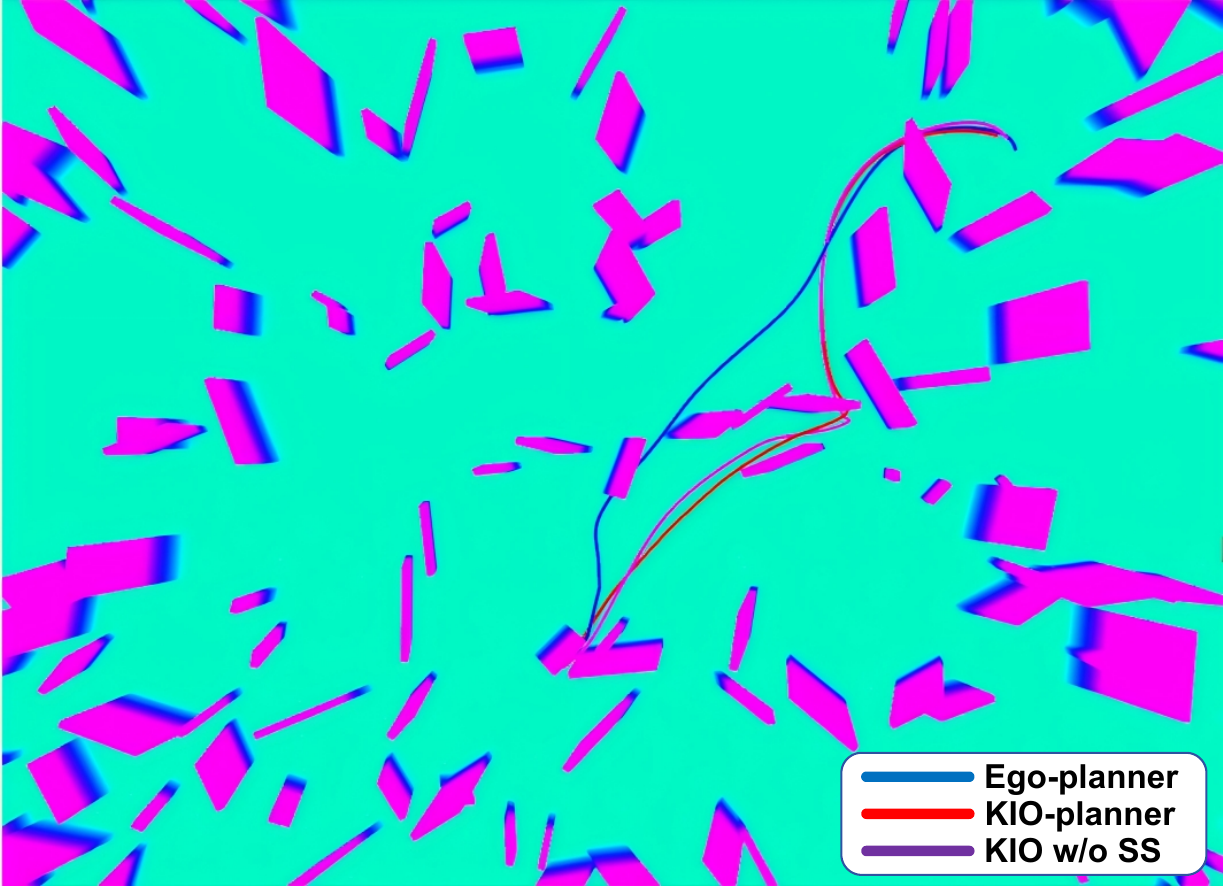}
    \caption{Qualitative 3D trajectory comparison in a highly confined wall environment at $v_{max} = 2.0$ m/s. The trajectories generated by EGO-planner, KIO-planner (w/o Safety Shield), and the complete KIO-planner are illustrated. As the speed limit increases, EGO-planner exhibits severe oscillations and cuts dangerously close to the corners, whereas KIO-planner consistently produces more centralized and smoother collision-free flight corridors.}
    \label{fig:traj_2}
\end{figure}

\begin{figure}[htbp]
    \centering
    \includegraphics[width=\columnwidth]{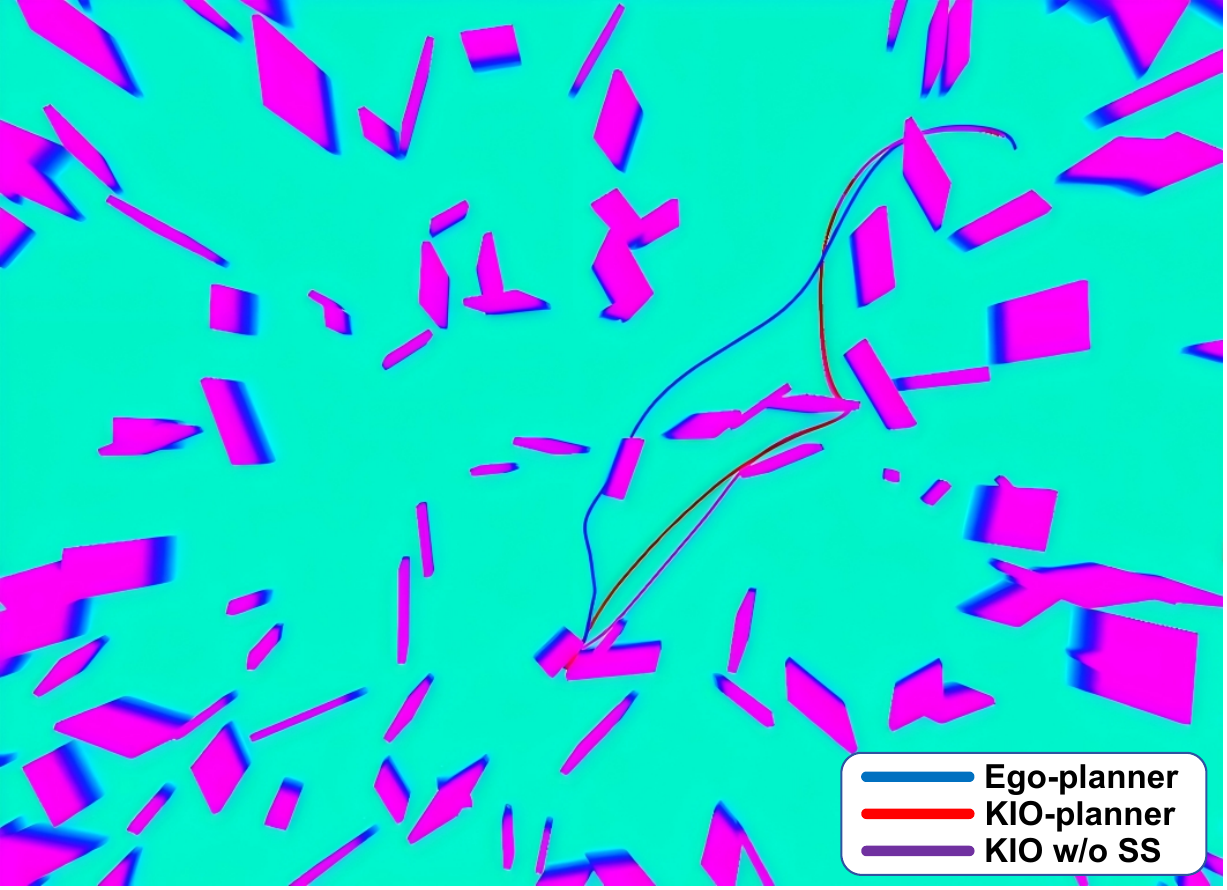}
    \caption{Trajectory comparison of the three algorithms at $v_{max} = 2.5$ m/s.}
    \label{fig:traj_25}
\end{figure}

\begin{figure}[htbp]
    \centering
    \includegraphics[width=\columnwidth]{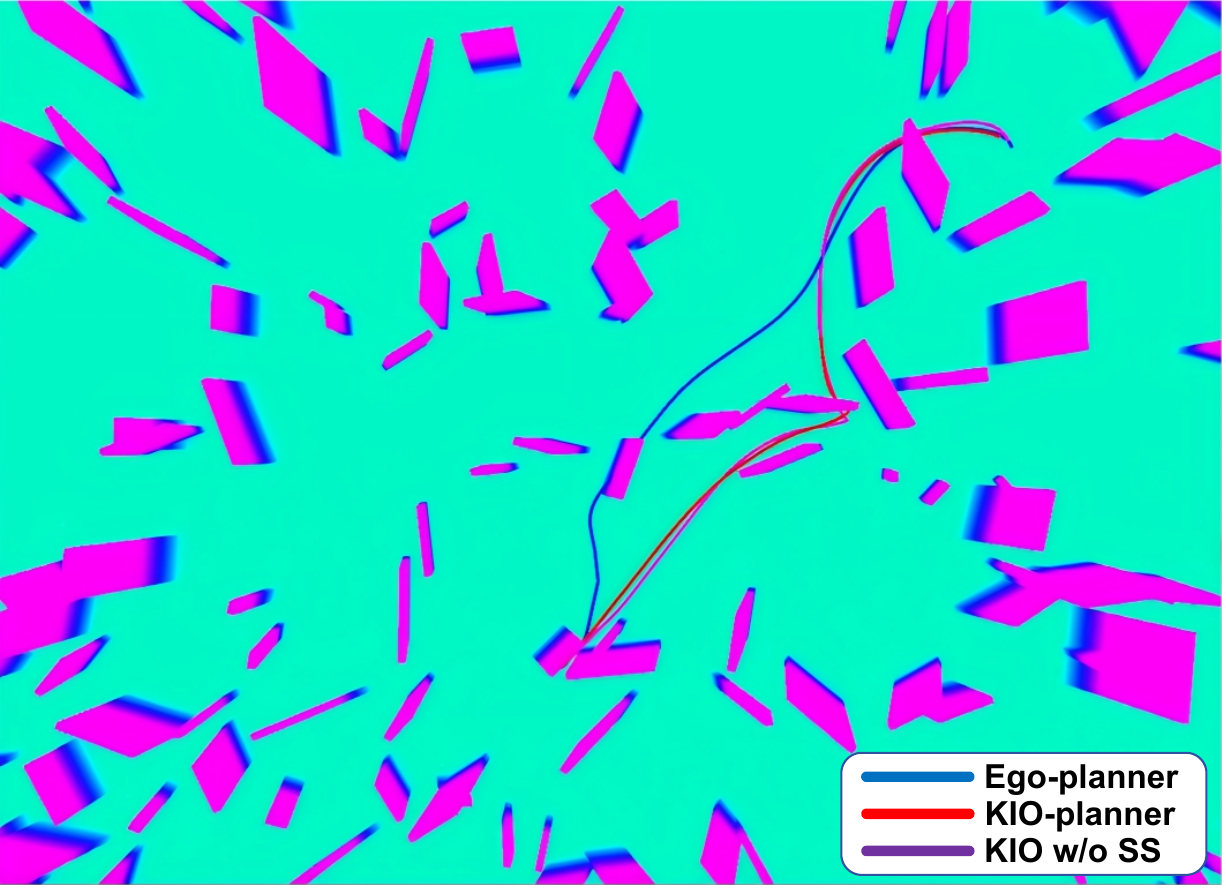}
    \caption{Trajectory comparison of the three algorithms at $v_{max} = 3.0$ m/s.}
    \label{fig:traj_3}
\end{figure}

To thoroughly evaluate the contribution of the Geometric Safety Shield and visualize the navigation behaviors, we present a comprehensive 3D trajectory comparison in Fig. \ref{fig:traj_2}, Fig. \ref{fig:traj_25} and Fig. \ref{fig:traj_3}. The figure illustrates the complete flight paths of EGO-planner, the ablation variant (\textit{KIO-planner (w/o Safety Shield)}), and the complete KIO-planner framework across three distinct speed limits ($v_{max} \in \{2.0, 2.5, 3.0\}$ m/s).

Quantitatively (Table \ref{tab:quantitative_full}), when the safety shield is disabled at the $3.0$ m/s limit, the minimum distance to obstacles drops from $0.76$ m to $0.65$ m, and the smoothness cost increases. This degradation confirms that the probabilistic nature of the raw neural network is insufficient for strict obstacle avoidance. 

Qualitatively, as vividly demonstrated in Fig. \ref{fig:traj_2}, Fig. \ref{fig:traj_25} and Fig. \ref{fig:traj_3}, EGO-planner consistently struggles with non-convex corners at all speed levels, frequently generating oscillatory paths and steering dangerously close to the walls. While the ablation variant (\textit{KIO-planner (w/o Safety Shield)}) shows improved spatial awareness due to the CBAM backbone, it still exhibits minor trajectory deviations near tight constraints. Conversely, the complete KIO-planner framework seamlessly navigates through the narrow gaps. The Dual Mapping mechanism acts as a deterministic hard constraint, proactively keeping the flight corridor highly centralized and strictly curtailing unsafe actions without compromising the aggressive flight speeds.

\section{CONCLUSION}
In this paper, we proposed KIO-planner, a novel attention-guided single-stage trajectory planning framework tailored for high-speed UAV navigation in highly confined and wall-dense environments. By integrating the Convolutional Block Attention Module (CBAM) into the perception backbone, KIO-planner effectively extracts fine-grained structural features and identifies narrow traversable gaps directly from raw depth images. To bridge the gap between learning-based predictions and strict safety requirements, we introduced a Dual Mapping mechanism. This mechanism explicitly enforces kinodynamic feasibility through physical bounds activation and provides a reliable collision-avoidance safeguard via a Geometric Safety Shield in the depth-pixel space.

Extensive high-fidelity simulated experiments demonstrated that KIO-planner achieves significantly higher safety margins, shorter flight paths, and smoother trajectories compared to the state-of-the-art EGO-planner, even at aggressive speeds up to 3.0 m/s. Specifically, our framework increased the minimum safety distance by 58.3\% while reducing control smoothness costs by over 28\%. Future work will focus on extending the Dual Mapping mechanism to handle dynamic obstacles and multi-UAV swarm coordination in highly unstructured environments.

\section{Acknowledgment}
The support provided by the Macao Science and Technology Development Fund (FDCT-MOST: 0018/2025/AMJ) and Macao Polytechnic University (MPU grant: RP/FCA-01/2025) enabled us to conduct the simulation experiments, data analysis, and manuscript preparation.

\end{document}